\pdfoutput=1

\documentclass[11pt,table]{article}

\usepackage[final]{acl}

\usepackage{times}
\usepackage{latexsym}

\usepackage[T1]{fontenc}

\usepackage[utf8]{inputenc}

\usepackage{microtype}

\usepackage{inconsolata}

\usepackage{graphicx}

\usepackage{booktabs}
\usepackage{multirow}
\usepackage{makecell}
\usepackage{array}
\usepackage{xcolor}
\usepackage{soul}
\usepackage{pifont}
\usepackage{amssymb}
\usepackage{amsmath}
\usepackage{todonotes}
\usepackage{titling}

\title{Spatial Coordinates as a Cell Language: \\ A Multi-Sentence Framework for Imaging Mass Cytometry Analysis}

\author{
 \textbf{Chi-Jane Chen\textsuperscript{*,1}}, 
 \textbf{Yuhang Chen\textsuperscript{*,1}},
 \textbf{Sukwon Yun\textsuperscript{*,1}},
\\
 \textbf{Natalie Stanley\textsuperscript{$\dagger$,1,2,3}},
 \textbf{Tianlong Chen\textsuperscript{$\dagger$,1}}
\\
 \textsuperscript{1}Department of Computer Science, The University of North Carolina at Chapel Hill,\\
 \textsuperscript{2}Computational Medicine Program, The University of North Carolina at Chapel Hill,\\
 \textsuperscript{3}Department of Genetics, The University of North Carolina at Chapel Hill\\
\\
}

\newcommand{\proposed}{\texttt{{Spatial2Sentence}}}

\begin{document}

\maketitle
\begin{abstract}
Image mass cytometry (IMC) enables high-dimensional spatial profiling by combining mass cytometry’s analytical power with spatial distributions of cell phenotypes. Recent studies leverage large language models (LLMs) to extract cell states by translating gene or protein expression into biological context. However, existing single-cell LLMs face two major challenges: (1) \textbf{Integration of spatial information}: they struggle to generalize spatial coordinates and effectively encode spatial context as text, and (2) \textbf{Treating each cell independently}: they overlook cell-cell interactions, limiting their ability to capture biological relationships. To address these limitations, we propose \proposed{}, a novel framework that integrates single-cell expression and spatial information into natural language using a multi-sentence approach. \proposed{} constructs expression similarity and distance matrices, pairing spatially adjacent and expressionally similar cells as positive pairs while using distant and dissimilar cells as negatives. These multi-sentence representations enable LLMs to learn cellular interactions in both expression and spatial contexts. Equipped with multi-task learning, \proposed{} outperforms existing single-cell LLMs on preprocessed IMC datasets, improving cell-type classification by \texttt{5.98\%} and clinical status prediction by \texttt{4.18\%} on the diabetes dataset while enhancing interpretability. The source code can be found here: \url{https://github.com/UNITES-Lab/Spatial2Sentence}.

\end{abstract}

\let\thefootnote\relax\footnotetext{\textsuperscript{*}Equal contribution. \quad \textsuperscript{$\dagger$} Co-corresponding authors. Correspondence: \texttt{\{natalies,tianlong\}@cs.unc.edu}}

\section{Introduction}

Single-cell technologies, such as flow and mass cytometry (CyTOF) and single-cell RNA sequencing, have revolutionized our ability to analyze cellular heterogeneity in blood and tissue samples \cite{bendall2012deep, brodin2019call, jagadeesh2022identifying}. These techniques provide high-resolution insights into the human immune system, enabling targeted therapeutic strategies for disease treatment and prevention \cite{reece2016microfluidic}. CyTOF, for instance, identifies immune cells based on protein expression \cite{bendall2012deep}. However, traditional suspension-based proteomic approaches lack spatial context, limiting our understanding of cell-cell interactions and tissue organization.

Recent advancements, such as imaging mass cytometry (IMC) and multiplexed ion beam imaging (MIBI), overcome this limitation by integrating mass cytometry with spatial profiling \cite{kakade2021using, shaaban2024cutting}. These next-generation single-cell proteomics technologies enable high-dimensional immune profiling while preserving spatial relationships \cite{giesen2014highly, nair2015mass}. Such insights are crucial for characterizing the tumor microenvironment, where spatial interactions between immune and tumor cells influence prognosis \cite{keren2018structured}. By capturing cell types, states, and interactions, spatial proteomics enhances our understanding of the immune system’s complexity and disease mechanisms \cite{hartmann2020immune}.

 Inspired by the advances in natural language processing (NLP), researchers have started conceptualizing cellular information as "words" and "sentences," enabling deep learning models to interpret cellular behavior within a context-dependent framework \cite{fang2024large}. A recent innovative approach applies large language model to extract cell (gene) state by translating cell (gene) expression information into biological context \cite{levine2023cell2sentence, chen2024genept}. By leveraging vast single-cell datasets, those models can generalize across gene expression dynamics, phenotype, and disease status, paving the way for new discoveries in immune system and therapeutic development. Despite their effectiveness in various downstream tasks, current approaches face two major bottlenecks:

\noindent \textbf{Integration of Spatial Information. } With rapid advancements in spatial transcriptomics and multiplexed imaging, capturing spatial context has become increasingly important~\cite{rao2021exploring, tian2023expanding, yun2024mew}. Cells interact within their microenvironment, influencing differentiation, immune response, and disease progression \cite{marx2021method}. However, current models rely solely on expression matrices (e.g., gene or protein) and overlook spatial organization, leading to suboptimal representations. Spatial context is essential for modeling intercellular interactions, uncovering cell-type-specific behaviors, and improving biological relevance. Ignoring it risks missing key regulatory mechanisms that shape cellular function within complex tissues.

 \noindent \textbf{Treating each cell independently. } Current single-cell LLM approaches~\cite{levine2023cell2sentence, cui2024scgpt} treat each cell as an independent entity, overlooking broader cellular interactions across similar or distinct groups of cells. Learning the similarities and distinctions between cells within the same or different functional groups (\textit{e.g.,} cell types or tissue niches) can enrich the biological context captured in cell sentences \cite{keren2018structured, hartmann2020immune}. Without this information, models may fail to capture key regulatory relationships or misinterpret cellular function within a tissue-specific context.
 


Driven by these motivations, we propose \proposed{}, a novel single-cell LLM framework that integrates spatial information into language using a multi-sentence approach to capture cellular interactions. Specifically, we construct an expression similarity matrix and a distance matrix from the expression matrix and spatial coordinates, respectively. For each cell, we identify the most similar and adjacent cells as positive pairs, while dissimilar and distant cells serve as negative pairs. These structured pairs are then used to prompt an LLM, enabling it to capture cell-cell interactions effectively. Using our newly preprocessed IMC dataset, which containing protein expression matrices, spatial coordinates, and designated cell-type annotations for diabetes and brain tumor samples, \proposed{} achieves state-of-the-art performance, improving cell-type classification by 5.98\% and clinical status prediction by 4.18\% compared to recent single-cell LLM approaches. Furthermore, our approach provides new insights into interpretability by identifying which cell types and protein markers are most crucial for distinguishing clinical states, offering a deeper understanding of their roles in disease progression.

  In summary, our main contributions are summarized as follows:

\begin{itemize}
    \item We highlight the limitations of existing single-cell LLMs, particularly their lack of spatial information integration and their limitation in capturing contextual information from neighboring cells, which can be important for improving annotation accuracy in spatial omics data.
    \item We propose a novel single-cell LLM framework, \proposed~that integrates spatial information as language and introduces multi-sentence contrastive prompting, enabling LLMs to capture cell interactions using positive and negative pairs based on both expression and spatial proximity.
    \item We preprocess and transform two IMC datasets into cell × protein feature matrices, providing spatial coordinates and cell-type annotations for each cell.
    \item Our method achieves state-of-the-art performance, surpassing previous models by 5.98\% in cell-type classification and 4.18\% in clinical status prediction on the Diabetes dataset.
\end{itemize}

\section{Related Works}
\noindent \textbf{Cells as language. }Transformer-based models have made significant advances in natural language processing due to their exceptional parallel processing capabilities and highly adaptable attention mechanisms. Building on these successes, researchers have begun applying transformer architectures to the modeling of single-cell data \cite{lan2024transformer}. scGPT is a deep learning-based approach designed for cell identification, particularly in the context of single-cell RNA sequencing (scRNA-seq) \cite{cui2024scgpt}. The model has demonstrated strong performance in various downstream tasks, including multi-batch integration, multiomic synthesis, cell-type classification, genetic perturbation prediction, and gene network inference. Cell2Sentence (C2S) is another pretrained model, fine-tuned from GPT-2, specifically designed to process textual sequences containing gene names \cite{levine2023cell2sentence}. The model generates new cell-level textual representations, and conversely, it can transform these textual sequences back into corresponding gene expression vectors. By converting cell text sequences back into gene expression profiles, C2S ranks genes based on their expression levels and preserves key information from the original data in the majority of cases. However, both scGPT and C2S have not yet been utilized for spatial datasets, such as those obtained from imaging mass cytometry (IMC) \cite{walsh2023decoding}. Additionally, if applied to spatial datasets without accounting for spatial location of individual cells, scGPT would overlook the geometric context inherent to imaging data, thereby forfeiting one of the key advantages provided by spatial modalities. Such existing approaches typically analyze cell data individually, failing to account for the potential interactions and relationships that may exist between pairs of cells.

    \begin{figure*}[!ht]
        \centering
        \includegraphics[width=1\linewidth]{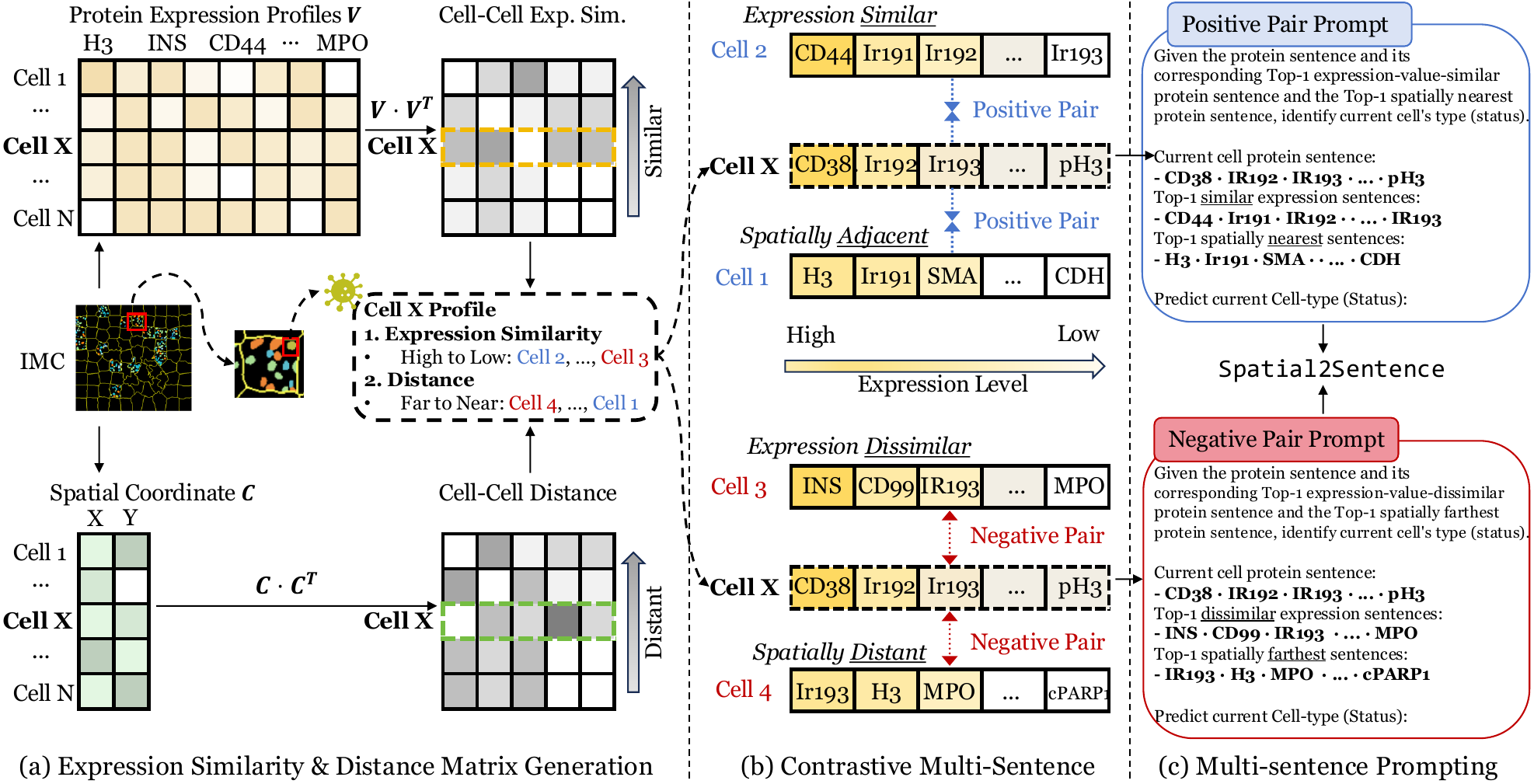}
        \vspace{-1em}
        \caption{Overall framework of~\proposed. (a) Given IMC data, we integrate both the protein expression matrix and the spatial coordinate information simultaneously. The protein expression matrix (with cells as rows and proteins as columns) is transposed (i.e., $V^T$) and multiplied to compute cosine similarity. Similarly, the spatial coordinate matrix (with columns representing the X and Y coordinates) is processed to obtain a distance matrix. For cell `X', we rank the cells based on their expression similarity (from highest to lowest) and their spatial proximity (from farthest to nearest). (b) Using these ranked cell indices, we perform Contrastive Multi-Sentence Generation. Here, the positive pairs consist of the top-$k$ cells in terms of both expression similarity and proximity to cell `X', while the negative cases use the top-$k$ cells with the most dissimilar expressions and that are most distant from cell `X'. (c) Finally, equipped with both positive and negative pairs, we prompt these pairs into LLMs to leverage their capability in handling proteins and spatial information within a multi-sentence framework that captures interactions among different cells. In the illustration, the top-$1$ case is used as an example; however, any $k$ can be used.}
        \label{fig:main_figure}
    \end{figure*}

\noindent \textbf{CyTOF. } Several methodologies have started leveraging CyTOF for cell annotation to study disease status or disease pathology \cite{milosevic2023different}. Take two examples of approaches that have been developed for performing cell annotation. Automated cell-type discovery and classification (ACDC) is an algorithm that can use in CyTOF or high-dimension mass cytometry like IMC to classify cell and new cell types by combining profile matching and semi-supervised learning \cite{lee2017automated}. Linear discriminant analysis (LDA) also another automatic classifier for cell classification enables the analysis of large CyTOF datasets without requiring prior biological knowledge of marker expression patterns across different cell types \cite{Abdelaal316034}. While these techniques enable automated annotate cells, they process each cell independently and disregard spatial information. Our study utilizes IMC data, which captures spatial information for each cell across all samples. Therefore, incorporating spatial information is crucial for a more comprehensive analysis. The integration of spatial characterization with cell expression remains largely unexplored, and effectively incorporating spatial information remains a challenge.

\section{Methodology}


\proposed~consists of three stages: expression similarity and distance matrix generation, contrastive multi-sentence learning, and multi-sentence prompting.  
In Section~\ref{preliminary}, we define the preliminaries and notations used in this paper.
In Section~\ref{matrix}, we describe the generation of expression similarity and distance matrices, enabling each cell to identify its similar counterparts based on expression and spatial information.  
In Section~\ref{contrastive_multi_sentence_prompting}, we introduce the multi-sentence technique, designed to capture interactions between cells through contrastive prompting with positive and negative pairs derived from the similarity and distance matrices.  
Figure~\ref{fig:main_figure} provides a detailed illustration of the overall framework of~\proposed.

\subsection{Preliminary Definitions}\label{preliminary}
Let $\mathbf{B} = \{b_1, b_2, \dots, b_M\}\in \mathbb{R}^{1 \times M}$ represent the names of the proteins, where each element $b_k$ corresponds to the name of the $k$-th protein. Let $ V \in \mathbb{R}^{N \times M} $ represent the protein expression profiles, where $ N $ is the number of cells and $ M $ is the number of proteins. Each entry $ v_{i,j} $ in the matrix $ V $ corresponds to the expression level of protein $ j $ in cell $ i $, and the spatial positions of the cells are denoted by $ C \in \mathbb{R}^{N \times 2} $, where each row $ \{x_i, y_i\} $ represents the 2D spatial coordinates of cell $ i $. 

Following C2S~\cite{levine2023cell2sentence}, we convert the expression data for each cell into a linguistic sentence. Specifically, each cell's protein expression profile $ \mathbf{v}_i = \{v_{i,1}, v_{i,2}, \dots, v_{i,M}\}$ is transformed into a sentence by rank-ordering the proteins according to their expression levels. This is based on the hypothesis that the rank of protein expression reflect the property of cell (\emph{i.e.}, certain cell types may exhibit higher expression levels of specific proteins). More formally, we convert the expression matrix $V$ into a cell sentence by ordering the proteins in decreasing order of expression:
\begin{equation}
r_{i,k} = \text{Rank}(v_{i,k}, \mathbf{v}_i)
\end{equation}
where $\text{Rank}(v_{i,k}, \mathbf{v}_i)$ denotes the descending rank of $v_{i,k}$ within $\mathbf{v}_i$. The resulting cell protein sentence is then given by:
\begin{equation}
S_{i} = \{b_{r_{i,1}},b_{r_{i,2}},\dots,b_{r_{i,M}}\}
\end{equation}
This transformation represents a cell's high-dimensional expression data as a token sequence, making it more accessible for LLMs.

\subsection{Expression Similarity \& Distance Matrix}\label{matrix}
To effectively incorporate both expression values and spatial information into the multi-sentence prompt design, we first compute two types of pairwise similarity measures: the cosine similarity based on the expression profiles and the Euclidean distance based on the spatial coordinates.

\noindent \textbf{Expression Similarity.}  
Let $ V \in \mathbb{R}^{N \times M} $ represent the matrix of protein expression profiles, where each row $ v_i $ is the expression vector of cell $i$ across $M$ proteins. The cosine similarity between two cells $i$ and $j$ is computed as:
\begin{equation}
\begin{split}
g_{i,j} = \text{CosSim}(i,j) = \frac{v_i \cdot v_j}{\|v_i\|_2 \|v_j\|_2}
\end{split}
\end{equation}
The resulting expression similarity matrix $ \mathbf{G} \in \mathbb{R}^{N \times N} $ captures the pairwise cosine similarities between the expression profiles of all cells, and each entry $ g_{i,j} $ represents the cosine similarity between cells $i$ and $j$.
\begin{equation}
\begin{split}
\mathbf{G} = \left[ \begin{matrix}
1 & g_{1,2} & \cdots & g_{1,N} \\
g_{2,1} & 1 & \cdots & g_{2,N} \\
\vdots & \vdots & \ddots & \vdots \\
g_{N,1} & g_{N,2} & \cdots & 1
\end{matrix} \right]
\end{split}
\end{equation}

\noindent \textbf{Spatial Proximity.}  
Given the matrix of spatial coordinates $ C \in \mathbb{R}^{N \times 2} $ where each row corresponds to the spatial coordinates $\{x_i, y_i\}$ of cell $i$, the Euclidean distance between cells $i$ and $j$ is
\begin{equation}
\begin{split}
d_{i,j} = \sqrt{(x_i - x_j)^2 + (y_i - y_j)^2}
\end{split}
\end{equation}
which is done pairwise for all cells, resulting in the distance matrix $ \mathbf{D} \in \mathbb{R}^{N \times N} $, where each entry $ D_{i,j} $ represents the Euclidean distance between the spatial coordinates of cells $i$ and $j$. We compute this for all pairs of cells, which output the full spatial distance matrix:
\begin{equation}
\begin{split}
\mathbf{D} = \left[ \begin{matrix}
0 & d_{1,2} & \cdots & d_{1,N} \\
d_{2,1} & 0 & \cdots & d_{2,N} \\
\vdots & \vdots & \ddots & \vdots \\
d_{N,1} & d_{N,2} & \cdots & 0
\end{matrix} \right]
\end{split}
\end{equation}

\noindent \textbf{Ranking Cells Sentences.}  
Once we compute $\mathbf{D}$ and $\mathbf{G}$, we rank the cells in terms of their proximity and similarity. For each cell $i$, we generate two ordered lists: \ding{172} Expression Similarity Ranked Sentences: This list contains the indices of cells ordered by their cosine similarity to cell $i$, indicating how similar their protein expression profiles are. \ding{173} Spatial Proximity Ranked Sentences: This list contains the indices of cells ordered by their Euclidean distance to cell $i$, indicating how physically close the cells are in the tissue.

\subsection{Contrastive Multi-sentence Generation \& Prompting}\label{contrastive_multi_sentence_prompting}

\noindent \textbf{Motivation.} Traditional single-cell approaches often represent each cell by a single sentence derived from its feature expression profile. While this method captures a cell's individual characteristics, it fails to account for the complex interrelationships between cells or provide a comprehensive understanding of cellular behavior. In particular, such representations overlook the potential nuances in how cells relate to each other in a biological context, especially in the presence of heterogeneous tissue environments. Therefore, relying on only one sentence per cell limits the model's ability to discern subtle differences or similarities between cells, which is crucial for tasks such as cell-type identification or disease status prediction.

\noindent \textbf{Multi-sentence Prompt.} To address this limitation, we introduce the concept of multi-sentence prompts, where each prompt includes the protein expression profiles of multiple cells. Instead of relying on a single sentence, the model processes pairs (or more) of sentences from different cells, which enhances its ability to capture both similarities and differences in protein expression across different cell types.

For two cells $i$ and $j$, each with their respective protein expression sentences $S_i$ and $S_j$, the multi-sentence prompt consists of:
\begin{equation}
\begin{split}
S_i = \{b_{r_{i,1}}, b_{r_{i,2}}, \dots, b_{r_{i,M}}\}, \\
S_j = \{b_{r_{j,1}}, b_{r_{j,2}}, \dots, b_{r_{j,M}}\}    
\end{split}
\end{equation}
Here, $r_{i,k}$ and $r_{j,k}$ denote the rank of the $k$-th protein in cell $i$ and cell $j$, respectively. 
This design encourages the LLMs to consider relationships between cells and can help identify commonalities or distinctions in cell types or disease status.


Once we have computed the spatial proximity and expression similarity for each cell, we design the multi-sentence contrastive prompting framework. This involves using both \texttt{positive-pair} and \texttt{negative-pair} prompts to provide the model with the necessary context for learning relationships between cells.

\noindent \texttt{\textbf{Positive-pair Prompt.}}  
In the positive-pair prompt, the goal is to guide the model to identify common characteristics between cells that share similar expression profiles and spatial proximities. We use cells that are close both in terms of expression similarity and spatial proximity. Specifically, for each cell $i$, we generate a positive-pair prompt by selecting the top $K$ most similar cells in terms of expression and the top $K$ spatially closest cells. The input prompt is formatted as \{\texttt{Prompt, $S_i$, $S_{\text{TopK}(D_i)}$, $S_{\text{TopK}(G_i)}$, Task (Predict cell type)}\} (See Figure~\ref{fig:main_figure} for details).
This prompt encourages the model to consider both molecular similarity and spatial proximity in making its predictions.

\noindent \texttt{\textbf{Negative-pair Prompt.}}  
For the negative-pair prompt, the objective is to contrast cells that exhibit dissimilar expression profiles and spatial distances. This allows the model to learn to distinguish between cells that are fundamentally different in terms of their biological properties. For each cell $i$, we generate a negative-pair prompt by selecting the top $K$ most dissimilar cells in terms of expression and the top $K$ spatially furthest cells. The negative-pair prompt is formatted as \{\texttt{Prompt,$S_i$,$S_{\text{TopK}(D_{\text{far}})}$,$S_{\text{TopK}(G_{\text{dissim}})}$,Task}\} ( See Figure~\ref{fig:main_figure} for details).
This prompt helps the model learn the distinctions between cells that are biologically or spatially divergent. With the positive-pair prompt and negative-pair prompt, the objective is to effectively distinguish between similar and dissimilar cells by leveraging both expression profiles and spatial contexts. The model is trained to predict cell types based on the information from the positive-pair and negative-pair prompts, enabling it to understand cellular behavior in the context of tissue heterogeneity.



\section{Experiments}

 \begin{figure*}[thbp]
    \centering
    \includegraphics[width=0.95\textwidth]{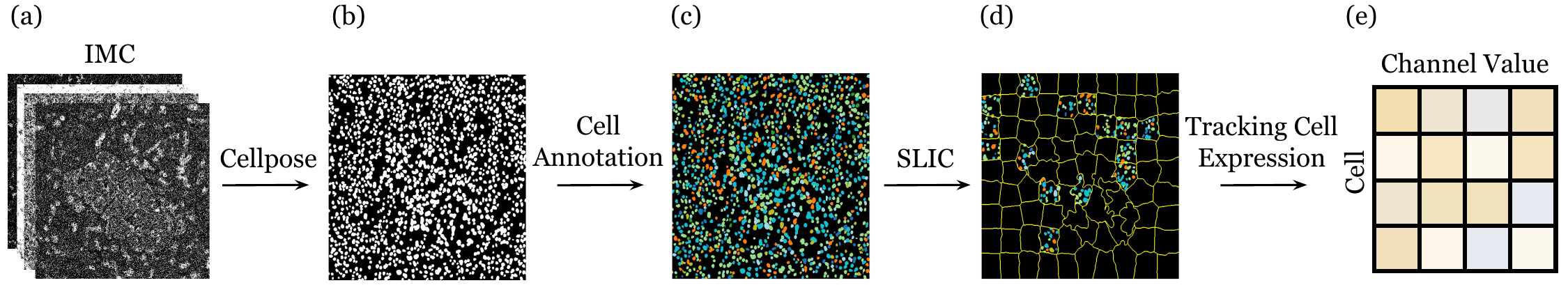}
    \vspace{-4pt}
    \caption{Given a ({\bf a}) multi-sample IMC dataset ({\bf b}), we used Cellpose to detect cell centers within superpixels ({\bf c}) and extracted cells from IMC images for cell-type annotation. ({\bf d}) We then applied the SLIC algorithm to segment images into superpixel region, ({\bf e}) generating a cell × protein feature matrix for analysis.}
    \label{dataset}
    \vspace{-2pt}
\end{figure*}

\begin{table*}[htbp]
    \centering
    \caption{Comparison of different models including scGPT, Geneformer, C2S, GenePT, scELMo, LangCell, and \proposed{} on diabetes and brain tumor datasets. Table shows classification accuracy (\%) across various settings, where \textit{Single-Task} (\textit{Multi-Task}) denotes single-tasking (multi-tasking), \textit{Type} represents cell-type classification, and \textit{Status} indicates clinical status prediction. Best results for each column are bolded.}
    \resizebox{0.9\linewidth}{!}{
    \begin{tabular}{l|cccc|cccc}
        \toprule
        \multirow{3}{*}{Model}
            & \multicolumn{4}{c|}{Diabetes}
            & \multicolumn{4}{c}{Brain Tumor} \\
        \cmidrule(lr){2-3} \cmidrule(lr){4-5} \cmidrule(lr){6-7} \cmidrule(lr){8-9}
             & \multicolumn{2}{c}{Single-Task} & \multicolumn{2}{c|}{Multi-Task}
             & \multicolumn{2}{c}{Single-Task} & \multicolumn{2}{c}{Multi-Task} \\
        \cmidrule(lr){2-5} \cmidrule(lr){6-9}
             & Type  & Status & Type & Status
             & Type & Status & Type & Status \\
        \midrule
        Geneformer & 31.62 & 55.78 & 34.29 & 62.02 & 48.10 & 54.09 & 50.14 & 58.16 \\
        GenePT & 37.33 & 60.15 & 39.98 & 64.02 & 51.90 & 52.20 & 53.67 & 58.31 \\
        scELMo & 34.33 & 56.14 & 37.95 & 62.03 & 49.44 & 50.26 & 51.66 & 55.27 \\
        LangCell & 40.83 & \textbf{64.64} & 41.17 & 72.51 & 53.38 & 55.70 & 55.16 & 63.80 \\
        scGPT & 32.45 & 58.12 & 34.50 & 65.38 & 47.92 & 56.27 & 50.11 & 62.04 \\
        scGPT w/ Spatial Info & 33.98 & 57.78 & 36.35 & 67.23 & 49.68 & 56.16 & 52.98 & 62.98 \\
        C2S & 36.37 & 60.45 & 37.54 & 72.55 & 49.03 & 52.06 & 51.86 & 54.04 \\
        C2S w/ Spatial Info & 36.98 & 62.15 & 38.03 & \textbf{74.11} & 51.12 & 53.93 & 53.22 & 56.09 \\
        \midrule
        \rowcolor{gray!20}
        \textbf{\proposed{} w/o Spatial Info} & 37.89 & 62.08 & 38.26 & 72.05 & 51.50 & 55.48 & 52.98 & \textbf{65.19} \\
        \rowcolor{gray!20}
        \textbf{\proposed{}} & \textbf{41.35} & 64.12 & \textbf{41.98} & 74.02 & \textbf{53.89} & \textbf{57.25} & \textbf{55.67} & 63.31 \\
        \bottomrule
    \end{tabular}
    }
    \vspace{-4pt}
    \vspace{-6pt}
    \label{tab:results}
\end{table*}

\subsection{Datasets}

To assess the model's ability to predict cell types and status, as well as to summarize cell-type abundances, we applied it to two multi-sample IMC image datasets, which are described below \cite{damond2019map,karimi2023single}. 

\noindent\textbf{Diabetes dataset.} This dataset profiles 67 diabetic and non-diabetic donors longitudinally from human pancreatic tissue. Among these, 33 samples are from non-diabetic donors, while 34 samples represent donors who developed long-term Type 1 diabetes. Long-term Type 1 diabetes refers to individuals who have lived with the disease for an extended period, typically several years after diagnosis. For those patients in the advanced stages of Type 1 diabetes are typically marked by prolonged autoimmune responses and extensive destruction of some particular cell-type, which lead to irreversible impairment of endogenous insulin production. Each cell in the dataset was characterized by 38 measured proteins. Moreover, we associated a clinical outcome with each sample in the dataset as \ding{172} long-duration diabetes or \ding{173} non-diabetic control. We annotated the cells into seven cell types, including T cell, Helper T cell, CD8 T/Cytotoxic T cell, Neutrophils, Monocytes/Macrophages, Immune cell, and other cell types.

\noindent\textbf{Brain tumor dataset.}  The glioblastoma IMC dataset includes samples from 118 glioblastoma patients, allowing detailed characterization of the tumor microenvironment (TME), and 46 brain metastasis (BrM) tumors from distinct patients.  A total of 21 protein markers were selected for analysis across both conditions. After excluding samples with missing information or that lack all common markers, we balanced the dataset to include samples from 37 glioblastoma donors and from 37 brain metastasis donors. Glioblastoma patients have a primary brain tumor that originates in the brain, typically from glial cells, and is known for being highly aggressive and fast-growing. In other condition, brain metastasis patients have secondary tumors that spread to the brain from other cancers in the body, such as lung or breast cancer. While both affect the brain, their origins, progression, and treatment strategies are fundamentally different. The images cover multiple tissue regions, including the tumor core, tumor margin, and tumor-adjacent normal tissue. Some samples contain multiple tissue regions, resulting in a final collection of 100 brain metastasis samples and 72 glioblastoma samples. For cell annotation, the cells were phenotyped into six distinct cell types, such as Tc cell, B cell, Astrocytes, M1-like MDMs, M2-like MDMs, and undefined cells.

For the data pre-processing, we utilized Cellpose to detect cell centers \cite{stringer2021cellpose}. After extracting cells from the IMC images, we performed downstream analyses, including cell clustering and cell-type annotation \cite{stanley2020vopo}. Cell-type annotation was manually defined at the cluster level rather than on a per-cell basis. Following cell detection, we employed the SLIC algorithm to segment the image into multiple super-pixel regions\cite{achanta2010slic}. This ultimately produced a matrix of cells $\times$ protein features within the defined superpixels for subsequent analysis.  The overview of data preprocessing shown in Fig. \ref{dataset}.

\begin{table}[t]
    \caption{\textbf{Ablation study} of model components, showing the accuracy of the model with different components removed on diabetes and brain tumor datasets.}
    \centering
    \resizebox{1.02\columnwidth}{!}{
    \begin{tabular}{l|cc|cc}
        \toprule
        \multirow{2}{*}{Component} & \multicolumn{2}{c|}{Diabetes} & \multicolumn{2}{c}{Brain Tumor} \\
        \cmidrule(lr){2-3} \cmidrule(lr){4-5}
             & Type & Status & Type & Status \\
        \midrule
        \rowcolor{gray!20}
        \proposed & \textbf{41.35} & \textbf{64.12} & \textbf{53.89} & \textbf{57.25} \\
        w/o Multi-sentence Prompting & 36.37 & 60.45 & 49.03 & 52.06 \\
        \midrule
        w/o Negative Pair & 38.78 & 62.12 & 51.87 & 56.35 \\
        w/o Positive Pair & 36.97 & 59.13 & 50.23 & 54.05 \\
        \midrule
        w/o Expression Similar Sentences & 39.86 & 60.24 & 52.12 & 52.14 \\
        w/o Spatial Proximal Sentences & 40.16 & 61.53 & 53.04 & 54.45 \\
        \midrule
        w/o Cosine Similarity (Random Select) & 36.02 & 62.13 & 49.45 & 53.82 \\
        w/o Euclidean Distance (Random Select) & 35.54 & 58.68 & 48.12 & 51.57 \\
        \midrule
    \end{tabular}
    }
    \vspace{-8pt}

    \label{tab:ablation}
    \vspace{-12pt}
\end{table}

\subsection{Experimental Details}
\noindent\textbf{Baselines.}
We compare our proposed method with several state-of-the-art and relevant models. Our comparisons include the widely-used large foundation model scGPT~\cite{cui2024scgpt}, a deep learning-based approach for cell-type identification in single-cell RNA sequencing; we also evaluate a variant, scGPT w/ Spatial Info, where spatial coordinates are explicitly provided as input. Additionally, we benchmark against Geneformer~\cite{lan2024transformer}, a transformer-based model for multi-omics data integration, and C2S~\cite{levine2023cell2sentence}, which generates cell-level textual representations from protein expression profiles. An extension, C2S w/ Spatial Info~\cite{walsh2023decoding}, which incorporates spatial data, is also considered. Furthermore, we include GenePT~\cite{chen2024genept}, an embedding model for single-cell biology focusing on feature-level representations; scELMo~\cite{liu2023scelmo}, which combines metadata and expression profiles using embeddings from language models; and LangCell~\cite{zhao2024langcell}, a pre-training framework integrating gene ranks with metadata for language-cell understanding. All these baseline models are evaluated on the diabetes and brain tumor datasets to assess their performance in cell-type classification and clinical status prediction tasks.

\begin{figure}[t]
    \centering
    \vspace{-6pt}
    \includegraphics[width=0.85\columnwidth]{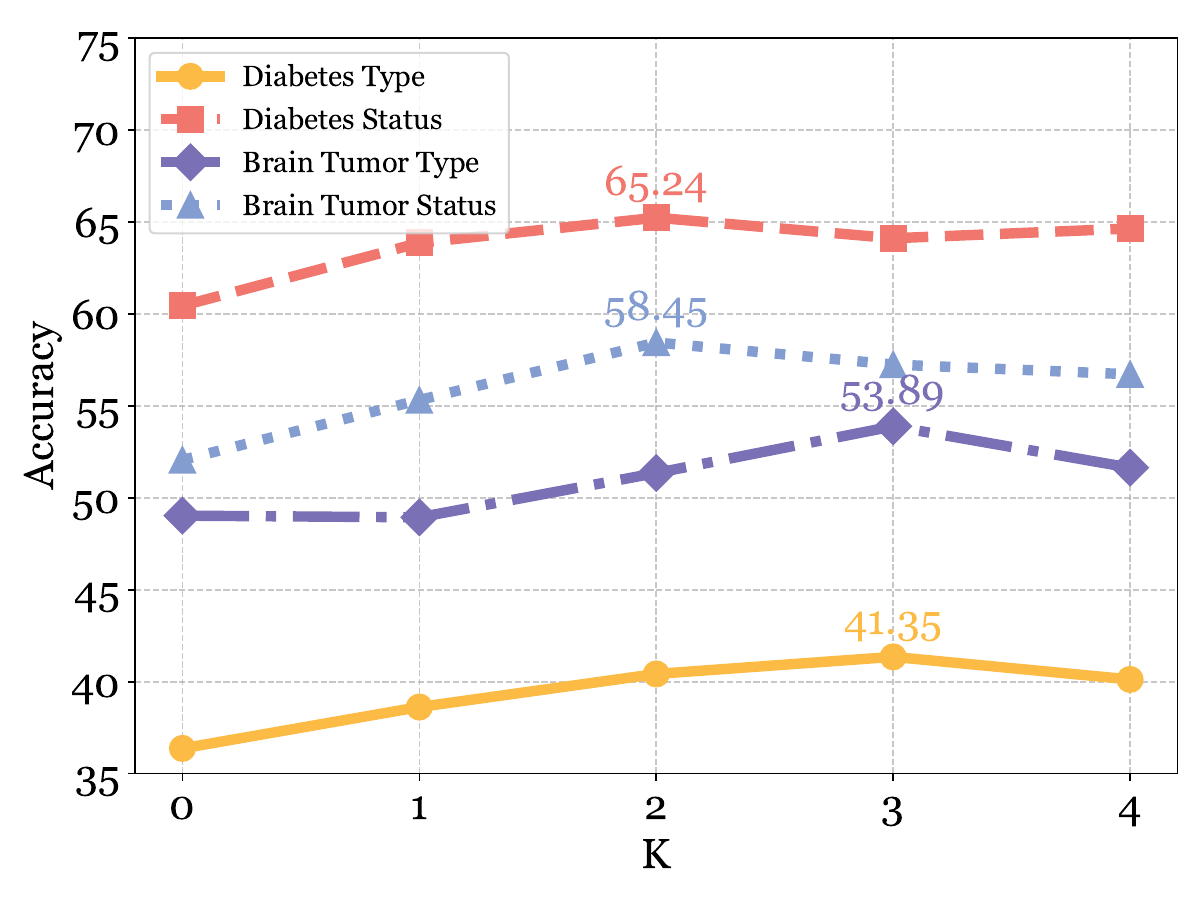}
    \vspace{-15pt}
    \caption{Accuracy performance of  \proposed{} across different values of the hyperparameter $K$.}
    \label{fig:K}
    \vspace{-8pt}
\end{figure}

\noindent\textbf{Tasks.} We perform both single-task and multi-task predictions to comprehensively assess our model's performance. Specifically, we evaluate cell Type and Status at the individual cell level. The performance metrics is classification accuracy.

\noindent\textbf{Experiment Details.}
We use the Llama-3.2-1B~\cite{dubey2024llama3} model for our experiments by default, fine-tuning it with the following training parameters: the batch size is set to 8 per device, and we apply a learning rate of 2e-4 with a cosine learning rate scheduler. The model is trained for 5 epochs, with a warm-up ratio of 0.05. The training is carried out using PyTorch on an NVIDIA RTX 6000 Ada Generation 48GB GPU.
The data set is divided into 90\% for training and 10\% for validation, with a separate test set reserved for the final evaluation. To ensure reproducibility, each experiment is repeated three times with different random seeds, and the results are averaged for reporting.

\begin{table*}[htbp]
    \caption{Classification accuracy comparison between the C2S method and \proposed{} across various LLMs on diabetes and brain tumor datasets. Single and Multi denote single-task and multi-task learning, respectively.}
    \vspace{-2mm}
    \centering
    \resizebox{0.85\textwidth}{!}{
    \begin{tabular}{l|l|cc|cc|cc|cc} 
        \toprule
        \multirow{3}{*}{Model} & \multirow{3}{*}{Method} 
            & \multicolumn{4}{c|}{Diabetes} 
            & \multicolumn{4}{c}{Brain Tumor} \\
        \cmidrule(lr){3-6} \cmidrule(lr){7-10}
             &  
             & \multicolumn{2}{c|}{Single} & \multicolumn{2}{c|}{Multi} 
             & \multicolumn{2}{c|}{Single} & \multicolumn{2}{c}{Multi} \\
        \cmidrule(lr){3-4} \cmidrule(lr){5-6} \cmidrule(lr){7-8} \cmidrule(lr){9-10}
             &  
             & Type  & Status & Type & Status 
             & Type & Status & Type & Status \\
        \midrule
        \multirow{2}{*}{GPT-2-Small} & C2S & 36.56 & 58.24 & 37.01 & 72.24 & 53.23 & 55.23 & 52.56 & 53.34 \\
                    & \cellcolor{gray!20}Ours & \cellcolor{gray!20}38.43 & \cellcolor{gray!20}59.70 & \cellcolor{gray!20}39.54 & \cellcolor{gray!20}71.40 & \cellcolor{gray!20}56.02 & \cellcolor{gray!20}57.26 & \cellcolor{gray!20}55.48 & \cellcolor{gray!20}52.78 \\
        \midrule
        \multirow{2}{*}{GPT-2-XL} & C2S & 37.13 & 62.34 & 37.81 & \bf{76.96} & 57.68 & 60.34 & 59.26 & 61.83 \\
        
                 & \cellcolor{gray!20}Ours & \cellcolor{gray!20}37.07 & \cellcolor{gray!20}62.72 & \cellcolor{gray!20}35.30 & \cellcolor{gray!20}74.83 & \cellcolor{gray!20}59.28 & \cellcolor{gray!20}61.21 & \cellcolor{gray!20}60.78 & \cellcolor{gray!20}\bf{63.51}  \\
        \midrule
        \multirow{2}{*}{Llama-2-7B} & C2S & 37.45 & 62.89 & 38.17 & 75.68 & 53.25 & 54.12 & 55.67 & 58.98 \\
                            & \cellcolor{gray!20}Ours & \cellcolor{gray!20}38.17 & \cellcolor{gray!20}59.72 & \cellcolor{gray!20}39.54 & \cellcolor{gray!20}74.40 & \cellcolor{gray!20}56.02 & \cellcolor{gray!20}57.26 & \cellcolor{gray!20}55.48 & \cellcolor{gray!20}56.78 \\
        \midrule
        \multirow{2}{*}{Llama-3.2-1B} & C2S & 36.37 & 60.45 & 37.54 & 72.55 & 49.03 & 52.06 & 51.86 & 54.04  \\
                           & \cellcolor{gray!20}Ours & \cellcolor{gray!20}\bf{41.35} & \cellcolor{gray!20}64.12 & \cellcolor{gray!20}\bf{41.98} & \cellcolor{gray!20}74.02 & \cellcolor{gray!20}53.89 & \cellcolor{gray!20}57.25 & \cellcolor{gray!20}55.67 & \cellcolor{gray!20}63.31 \\
        \midrule
        \multirow{2}{*}{Gemma-2-9B} & C2S & 38.34 & 65.12 & 38.01 & 76.78 & 59.53 & 57.39 & 60.32 & 59.32 \\
                            & \cellcolor{gray!20}Ours & \cellcolor{gray!20}40.45 & \cellcolor{gray!20}\bf{67.24} & \cellcolor{gray!20}40.31 & \cellcolor{gray!20}75.89 & \cellcolor{gray!20}\bf{62.13} & \cellcolor{gray!20}\bf{63.23} &  \cellcolor{gray!20}\bf{62.45} & \cellcolor{gray!20}61.23 \\ \bottomrule
    \end{tabular}
    }
    \label{tab:llm}
\end{table*}

\begin{table}[ht]
    \caption{\textbf{Sensitivity} on different spatial distances.}
    \vspace{-1mm}
    \centering
    \resizebox{\columnwidth}{!}{
    \begin{tabular}{c|cc|cc}
        \toprule
        \multirow{2}{*}{Component} & \multicolumn{2}{c|}{Diabetes} & \multicolumn{2}{c}{Brain Tumor} \\
        \cmidrule(lr){2-3} \cmidrule(lr){4-5}
             & Type & Status & Type & Status \\
        \midrule
        L1 Norm      & \bf{41.35} & 63.11 & 53.19 & 56.03 \\
        Cosine Distance & 38.25 & 62.24 & 50.56 & 53.31 \\
        \rowcolor{gray!20} Euclidean Distance & \bf{41.35} & \bf{64.12} & \bf{53.89} & \bf{57.25} \\
        \bottomrule
    \end{tabular}
    }
    \vspace{-4pt}

    \label{tab:spatial_sensitivity}
\end{table}

\begin{table}[ht]
    \caption{\textbf{Sensitivity} on different expression similarities.}
    \vspace{-1mm}
    \centering
    \resizebox{\columnwidth}{!}{
    \begin{tabular}{c|cc|cc}
        \toprule
        \multirow{2}{*}{Component} & \multicolumn{2}{c|}{Diabetes} & \multicolumn{2}{c}{Brain Tumor} \\
        \cmidrule(lr){2-3} \cmidrule(lr){4-5}
             & Type & Status & Type & Status \\
        \midrule
        Pearson Correlation & 39.16 & \bf{65.56} & 49.76 & 56.86 \\
        Euclidean Distance  & 39.22 & 63.87 & 48.94 & 54.32 \\
        \rowcolor{gray!20} Cosine Similarity   & \bf{41.35} & 64.12 & \bf{53.89} & \bf{57.25} \\
        \bottomrule
    \end{tabular}
    }
    \vspace{-4pt}
    \label{tab:expression_sensitivity}
    \vspace{-8pt}
\end{table}

\subsection{Primary Results}

\noindent\textbf{Comparison to State-of-the-Art.}
Table~\ref{tab:results} provides a detailed comparison of our method with several state-of-the-art and relevant models, now including scGPT with added spatial information, GenePT~\cite{chen2024genept}, scELMo~\cite{liu2023scelmo}, and LangCell~\cite{zhao2024langcell}, Geneformer~\cite{theodoris2023transfer},  alongside the baselines like scGPT~\cite{cui2024scgpt}, and C2S~\cite{levine2023cell2sentence} with and without spatial information. While the performance landscape varies across different tasks and datasets, \proposed{} demonstrates leading results in several key areas, particularly in cell-type classification tasks and certain clinical status prediction scenarios, underscoring the benefits of its spatial-aware multi-sentence framework (further details in Table~\ref{tab:results}). For instance, \proposed{} achieves top performance in Diabetes single-task cell-type classification (41.35\%) and Brain Tumor single-task cell-type classification (53.89\%). The inclusion of spatial information within the \proposed{} framework generally enhances its ability to capture relevant patterns effectively.

\noindent\textbf{Effect of Multi-task learning.} The results in Table~\ref{tab:results} demonstrate the effectiveness of multi-task learning in improving model performance across both the diabetes and brain tumor datasets. For instance, on the diabetes dataset, the multi-task setting improves cell-type prediction accuracy by 3.44\% and status prediction accuracy by 0.91\% over the single-task approach. 
We argue that this performance boost can be attributed to the model's ability to learn shared representations across related tasks, enhancing its generalization capability. Furthermore, patient-level status information can aid in improving cell-type prediction, as the model learns broader context about disease states. Similarly, knowing the cell types contributes to better status prediction, as it provides crucial biological insights into the patient's condition, reinforcing the interdependence between these tasks.

\begin{figure}[t]
    \centering
    \includegraphics[width=\columnwidth]{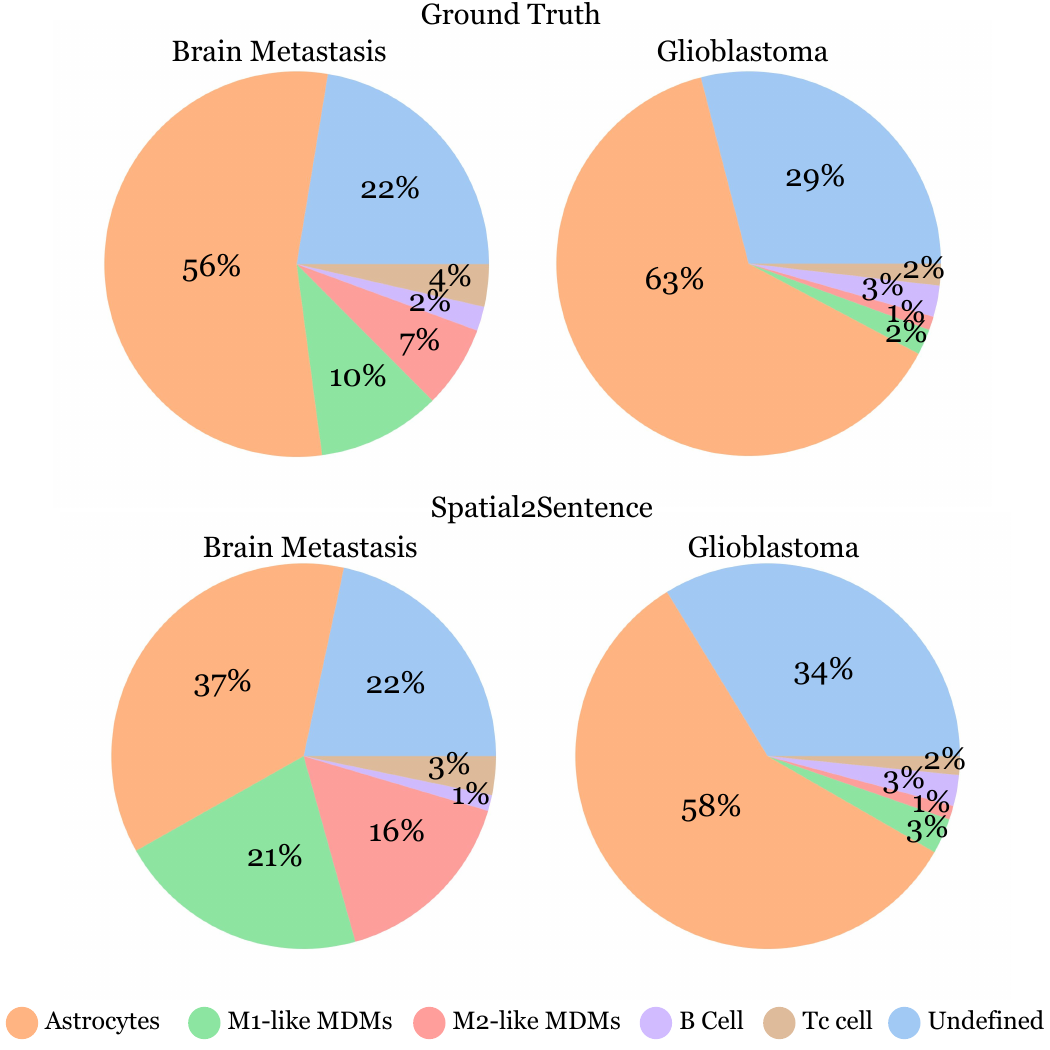}
    \vspace{-4mm}
    \caption{Cell-type distribution in brain tumor dataset between ground truth and \proposed{} result.}
    \label{pie_chart}
    \vspace{-3mm}
\end{figure}

\noindent\textbf{Cell-Type Frequency Analysis.} To summarize the cell type distribution in different disease statuses, we conduct experiment on brain dataset shown in Fig. \ref{pie_chart}. The finding from Spatial2Sentence indicates a sharp increase in M1-like MDMs in brain metastasis patients, which aligns with the trend observed in the ground truth pie chart. Additionally, previous studies have reported that brain metastases appear in higher proportion of M1-like monocyte-derived macrophages (MDMs) which further supports our finding \cite{karimi2023single, schreurs2025immune}.

\begin{table*}[!th]
\caption{Ablation study on contrastive negative pair selection strategies for \proposed{}. }
\label{tab:ablation_negative_pair_strategy}
\centering
\resizebox{0.75\textwidth}{!}{%
\begin{tabular}{l|cc|cc}
\toprule
\multirow{2}{*}{Negative Pair Size} & \multicolumn{2}{c|}{Diabetes (Multi-Task)} & \multicolumn{2}{c}{Brain Tumor (Multi-Task)} \\
\cmidrule(lr){2-3} \cmidrule(lr){4-5}
& Type & Status & Type & Status \\
\midrule
Top-1 & 41.08 & 73.72 & 54.25 & 62.78 \\
\rowcolor{gray!20} \textbf{Top 1-3 } & \textbf{41.98} & \textbf{74.02} & \textbf{55.67} & \textbf{63.31} \\
Top 4-6 & 40.42 & 73.15 & 54.93 & 63.64 \\
Top 7-9  & 39.09 & 73.37 & 53.56 & 62.18 \\
\bottomrule
\end{tabular}%
}
\end{table*}

\subsection{Diagnostic Analysis}\label{sec:diagno}
\noindent\textbf{Ablation Study.} In parallel, we conduct ablation experiments to assess the impact of various model components on the overall performance. 
The results in Table~\ref{tab:ablation} shows that \ding{172} removing multi-sentence prompting decreases contextual understanding, while excluding either pair type (positive or negative) reduces ability to differentiate cells, with positive pairs being more critical. \ding{173} Removing expression similarity harms performance more than spatial proximity, highlighting the importance of molecular features. \ding{174} Replacing structured similarity with random selection leads to a significant performance drop, emphasizing the importance of these measures.


\noindent\textbf{Sensitive Study.}
\ding{172} \ul{Hyperparameter $K$} determines amount of cell sentences for generating the positive-pair and negative-pair multi-sentence prompts. We evaluated \proposed{} across various $K$ values in Figure \ref{fig:K}. \proposed{} ($K\geq1$) consistently ourperforms baseline ($K=0$). \ding{173} Sensitivity studies on \ul{spatial distance methods} (Table \ref{tab:spatial_sensitivity}) and \ul{expression similarity methods} (Table \ref{tab:expression_sensitivity}) show that Euclidean Distance and Cosine Similarity yield the best results.

\noindent\textbf{Performance Across Different LLMs.} We evaluate the performance of our method across different LLMs, including GPT-2-Small~\cite{gpt2radford2019language}, GPT-2-XL~\cite{gpt2radford2019language}, Llama-2-7B~\cite{touvron2023llama}, Llama-3.2-1B~\cite{dubey2024llama3}, and Gemma-2-9B~\cite{team2024gemma}. Table~\ref{tab:llm} presents the classification accuracy of both the C2S model and our proposed method under each LLM configuration on the diabetes and brain tumor datasets. We find that \ding{172} Our method outperforms C2S in cell-type classification and disease status prediction for most LLM configurations. \ding{173} On the diabetes dataset, our method achieves 41.35\% accuracy in cell-type classification using Llama-3.2-1B, compared to C2S’s 36.37\%, but some models like GPT-2-Small show less improvement. We argue this is because smaller models like GPT-2-Small have relatively limited capacity to handle longer prompts.

\noindent\textbf{Impact of Negative Pair Selection Strategy.}
The construction of informative negative pairs is crucial for contrastive learning. Our framework selects the top-$K$ most dissimilar cells (in terms of expression and spatial distance) to form negative pairs, rather than relying on random sampling or a single extreme case. To evaluate the impact of this selection, we performed an ablation study by varying the criteria for dissimilar cells. Table~\ref{tab:ablation_negative_pair_strategy} shows the performance of \proposed{} when using different top-$K$ ranges for selecting negative cells for the multi-sentence prompts. The results indicate that using the top 1-3 most dissimilar and distant cells provides the most effective contrastive signal, leading to the best performance in multi-task settings. This suggests that carefully chosen, highly contrasting negative examples are more beneficial than moderately or extremely dissimilar ones beyond a certain threshold.

\section{Conclusion}

We proposed \proposed{}, a novel framework that jointly integrates spatial information and cell expression by modeling cells as sentences. \proposed{} effectively captures cellular similarities and distinctions across diverse cell types, enhancing the biological context encoded in language. Experiments on two IMC datasets show that \proposed{} consistently outperforms existing methods. By introducing a multi-sentence contrastive prompting strategy that leverages both spatial proximity and cell-level similarity, we demonstrate that even lightweight LLMs can gain substantial performance boosts from spatially informed prompts. These findings suggest that scaling to stronger models could further advance this direction, offering a promising bridge between spatial biology and large language models.



\newpage

\section*{Limitations and Future Works}
However, there are still substantial opportunities to expand this methodology to multi-model framework, incorporating multi-omic data such as genomics, transcriptomics, or proteomics. A full biological view of cell heterogeneity and disease mechanisms. Additionally, enhancing our model into large LLMs, e.g., with 70B-80B parameters will further improve its ability to capture more complex biological insights. Furthermore, we plan to explore robust learning strategies for scenarios when some modalities are missing, ensuring the model remains effective even given incomplete data. By addressing these challenges, we aim to develop flexible and powerful framework for biology analysis.

\section*{Ethical Statement}
To the best of our knowledge, the Diabetes and Brain datasets used in this study have been compiled from publicly available sources, ensuring compliance with ethical guidelines and avoiding the inclusion of sensitive or private information. The primary focus of Spatial2Sentence is to enhance the representation of single-cell expression and spatial interactions by leveraging a multi-sentence natural language approach. This method is specifically designed for biomedical research, distinguishing it from general-purpose language models used in dialogue systems. However, we acknowledge potential ethical concerns, including biases in dataset representation and the risk of misinterpretation of biological findings. While our framework inherently reduces the likelihood of generating harmful content, we emphasize the need for responsible usage and validation to prevent potential misuse, particularly in clinical or diagnostic applications.

\section*{Acknowledgments}
This research was, in part, funded by the National Institutes of Health (NIH) under other transactions 1OT2OD03804501. The views and conclusions contained in this document are those of the authors and should not be interpreted as representing official policies, either expressed or implied, of the NIH. This work was also supported by the National Institute of Aging of the NIH award number 5R21AG084251-02 (NS) and the National Institute of Allergy and Infectious Diseases of the NIH award number 5R21AI171745-02 (NS).
\bibliography{custom}

\begin{thebibliography}{36}
\providecommand{\natexlab}[1]{#1}

\bibitem[{Abdelaal et~al.(2018)Abdelaal, van Unen, Höllt, Koning, Reinders, and Mahfouz}]{Abdelaal316034}
Tamim Abdelaal, Vincent van Unen, Thomas Höllt, Frits Koning, Marcel~J.T. Reinders, and Ahmed Mahfouz. 2018.
\newblock Predicting cell types in single cell mass cytometry data.
\newblock \emph{bioRxiv}, 19(6):759--769.

\bibitem[{Achanta et~al.(2010)Achanta, Shaji, Smith, Lucchi, Fua, and S{\"u}sstrunk}]{achanta2010slic}
Radhakrishna Achanta, Appu Shaji, Kevin Smith, Aurelien Lucchi, Pascal Fua, and Sabine S{\"u}sstrunk. 2010.
\newblock Slic superpixels.

\bibitem[{Bendall et~al.(2012)Bendall, Nolan, Roederer, and Chattopadhyay}]{bendall2012deep}
Sean~C Bendall, Garry~P Nolan, Mario Roederer, and Pratip~K Chattopadhyay. 2012.
\newblock A deep profiler's guide to cytometry.
\newblock \emph{Trends in immunology}, 33(7):323--332.

\bibitem[{Brodin et~al.(2019)Brodin, Duffy, and Quintana-Murci}]{brodin2019call}
Petter Brodin, Darragh Duffy, and Lluis Quintana-Murci. 2019.
\newblock A call for blood—in human immunology.
\newblock \emph{Immunity}, 50(6):1335--1336.

\bibitem[{Chen and Zou(2024)}]{chen2024genept}
Yiqun Chen and James Zou. 2024.
\newblock Genept: a simple but effective foundation model for genes and cells built from chatgpt.
\newblock \emph{bioRxiv}, pages 2023--10.

\bibitem[{Cui et~al.(2024)Cui, Wang, Maan, Pang, Luo, Duan, and Wang}]{cui2024scgpt}
Haotian Cui, Chloe Wang, Hassaan Maan, Kuan Pang, Fengning Luo, Nan Duan, and Bo~Wang. 2024.
\newblock scgpt: toward building a foundation model for single-cell multi-omics using generative ai.
\newblock \emph{Nature Methods}, pages 1--11.

\bibitem[{Damond et~al.(2019)Damond, Engler, Zanotelli, Schapiro, Wasserfall, Kusmartseva, Nick, Thorel, Herrera, Atkinson et~al.}]{damond2019map}
Nicolas Damond, Stefanie Engler, Vito~RT Zanotelli, Denis Schapiro, Clive~H Wasserfall, Irina Kusmartseva, Harry~S Nick, Fabrizio Thorel, Pedro~L Herrera, Mark~A Atkinson, et~al. 2019.
\newblock A map of human type 1 diabetes progression by imaging mass cytometry.
\newblock \emph{Cell metabolism}, 29(3):755--768.

\bibitem[{Dubey et~al.(2024)Dubey, Jauhri, Pandey, Kadian, Al-Dahle, Letman, Mathur, Schelten, Yang, Fan et~al.}]{dubey2024llama3}
Abhimanyu Dubey, Abhinav Jauhri, Abhinav Pandey, Abhishek Kadian, Ahmad Al-Dahle, Aiesha Letman, Akhil Mathur, Alan Schelten, Amy Yang, Angela Fan, et~al. 2024.
\newblock The llama 3 herd of models.
\newblock \emph{arXiv preprint arXiv:2407.21783}.

\bibitem[{Fang et~al.(2024)Fang, Wang, Song, Long, Lu, Chen, Feng, Zhou, and Li}]{fang2024large}
Chen Fang, Yidong Wang, Yunze Song, Qingqing Long, Wang Lu, Linghui Chen, Guihai Feng, Yuanchun Zhou, and Xin Li. 2024.
\newblock How do large language models understand genes and cells.
\newblock \emph{ACM Transactions on Intelligent Systems and Technology}.

\bibitem[{Giesen et~al.(2014)Giesen, Wang, Schapiro, Zivanovic, Jacobs, Hattendorf, Sch{\"u}ffler, Grolimund, Buhmann, Brandt et~al.}]{giesen2014highly}
Charlotte Giesen, Hao~AO Wang, Denis Schapiro, Nevena Zivanovic, Andrea Jacobs, Bodo Hattendorf, Peter~J Sch{\"u}ffler, Daniel Grolimund, Joachim~M Buhmann, Simone Brandt, et~al. 2014.
\newblock Highly multiplexed imaging of tumor tissues with subcellular resolution by mass cytometry.
\newblock \emph{Nature methods}, 11(4):417--422.

\bibitem[{Hartmann and Bendall(2020)}]{hartmann2020immune}
Felix~J Hartmann and Sean~C Bendall. 2020.
\newblock Immune monitoring using mass cytometry and related high-dimensional imaging approaches.
\newblock \emph{Nature Reviews Rheumatology}, 16(2):87--99.

\bibitem[{Jagadeesh et~al.(2022)Jagadeesh, Dey, Montoro, Mohan, Gazal, Engreitz, Xavier, Price, and Regev}]{jagadeesh2022identifying}
Karthik~A Jagadeesh, Kushal~K Dey, Daniel~T Montoro, Rahul Mohan, Steven Gazal, Jesse~M Engreitz, Ramnik~J Xavier, Alkes~L Price, and Aviv Regev. 2022.
\newblock Identifying disease-critical cell types and cellular processes by integrating single-cell rna-sequencing and human genetics.
\newblock \emph{Nature genetics}, 54(10):1479--1492.

\bibitem[{Kakade et~al.(2021)Kakade, Weiss, and Cantley}]{kakade2021using}
Vijayakumar~R Kakade, Marlene Weiss, and Lloyd~G Cantley. 2021.
\newblock Using imaging mass cytometry to define cell identities and interactions in human tissues.
\newblock \emph{Frontiers in Physiology}, 12:817181.

\bibitem[{Karimi et~al.(2023)Karimi, Yu, Maritan, Perus, Rezanejad, Sorin, Dankner, Fallah, Dor{\'e}, Zuo et~al.}]{karimi2023single}
Elham Karimi, Miranda~W Yu, Sarah~M Maritan, Lucas~JM Perus, Morteza Rezanejad, Mark Sorin, Matthew Dankner, Parvaneh Fallah, Samuel Dor{\'e}, Dongmei Zuo, et~al. 2023.
\newblock Single-cell spatial immune landscapes of primary and metastatic brain tumours.
\newblock \emph{Nature}, 614(7948):555--563.

\bibitem[{Keren et~al.(2018)Keren, Bosse, Marquez, Angoshtari, Jain, Varma, Yang, Kurian, Van~Valen, West et~al.}]{keren2018structured}
Leeat Keren, Marc Bosse, Diana Marquez, Roshan Angoshtari, Samir Jain, Sushama Varma, Soo-Ryum Yang, Allison Kurian, David Van~Valen, Robert West, et~al. 2018.
\newblock A structured tumor-immune microenvironment in triple negative breast cancer revealed by multiplexed ion beam imaging.
\newblock \emph{Cell}, 174(6):1373--1387.

\bibitem[{Lan et~al.(2024)Lan, He, Liu, Chen, Cao, and Peng}]{lan2024transformer}
Wei Lan, Guohang He, Mingyang Liu, Qingfeng Chen, Junyue Cao, and Wei Peng. 2024.
\newblock Transformer-based single-cell language model: A survey.
\newblock \emph{Big Data Mining and Analytics}, 7(4):1169--1186.

\bibitem[{Lee et~al.(2017)Lee, Kosoy, Becker, Dudley, and Kidd}]{lee2017automated}
Hao-Chih Lee, Roman Kosoy, Christine~E Becker, Joel~T Dudley, and Brian~A Kidd. 2017.
\newblock Automated cell type discovery and classification through knowledge transfer.
\newblock \emph{Bioinformatics}, 33(11):1689--1695.

\bibitem[{Levine et~al.(2023)Levine, Rizvi, L{\'e}vy, Pallikkavaliyaveetil, Zhang, Chen, Ghadermarzi, Wu, Zheng, Vrkic et~al.}]{levine2023cell2sentence}
Daniel Levine, Syed~Asad Rizvi, Sacha L{\'e}vy, Nazreen Pallikkavaliyaveetil, David Zhang, Xingyu Chen, Sina Ghadermarzi, Ruiming Wu, Zihe Zheng, Ivan Vrkic, et~al. 2023.
\newblock Cell2sentence: teaching large language models the language of biology.
\newblock \emph{BioRxiv}, pages 2023--09.

\bibitem[{Liu et~al.(2023)Liu, Chen, Zheng, Luo, and Zhao}]{liu2023scelmo}
Tianyu Liu, Tianqi Chen, Wangjie Zheng, Xiao Luo, and Hongyu Zhao. 2023.
\newblock scelmo: Embeddings from language models are good learners for single-cell data analysis.
\newblock \emph{bioRxiv}, pages 2023--12.

\bibitem[{Marx(2021)}]{marx2021method}
Vivien Marx. 2021.
\newblock Method of the year: spatially resolved transcriptomics.
\newblock \emph{Nature methods}, 18(1):9--14.

\bibitem[{Milosevic(2023)}]{milosevic2023different}
Vladan Milosevic. 2023.
\newblock Different approaches to imaging mass cytometry data analysis.
\newblock \emph{Bioinformatics Advances}, 3(1):vbad046.

\bibitem[{Nair et~al.(2015)Nair, Mei, Chen, Hale, Nolan, Maecker, Genovese, Fathman, and Whiting}]{nair2015mass}
Nitya Nair, Henrik~E Mei, Shih-Yu Chen, Matthew Hale, Garry~P Nolan, Holden~T Maecker, Mark Genovese, C~Garrison Fathman, and Chan~C Whiting. 2015.
\newblock Mass cytometry as a platform for the discovery of cellular biomarkers to guide effective rheumatic disease therapy.
\newblock \emph{Arthritis research \& therapy}, 17:1--9.

\bibitem[{Radford et~al.(2019)Radford, Wu, Child, Luan, Amodei, Sutskever et~al.}]{gpt2radford2019language}
Alec Radford, Jeffrey Wu, Rewon Child, David Luan, Dario Amodei, Ilya Sutskever, et~al. 2019.
\newblock Language models are unsupervised multitask learners.
\newblock \emph{OpenAI blog}, 1(8):9.

\bibitem[{Rao et~al.(2021)Rao, Barkley, Fran{\c{c}}a, and Yanai}]{rao2021exploring}
Anjali Rao, Dalia Barkley, Gustavo~S Fran{\c{c}}a, and Itai Yanai. 2021.
\newblock Exploring tissue architecture using spatial transcriptomics.
\newblock \emph{Nature}, 596(7871):211--220.

\bibitem[{Reece et~al.(2016)Reece, Xia, Jiang, Noren, McBride, and Oakey}]{reece2016microfluidic}
Amy Reece, Bingzhao Xia, Zhongliang Jiang, Benjamin Noren, Ralph McBride, and John Oakey. 2016.
\newblock Microfluidic techniques for high throughput single cell analysis.
\newblock \emph{Current opinion in biotechnology}, 40:90--96.

\bibitem[{Schreurs et~al.(2025)Schreurs, Vom~Stein, J{\"u}nger, Timmer, Noh, Buettner, Kashkar, Neuschmelting, Goldbrunner, and Nguyen}]{schreurs2025immune}
Luca~D Schreurs, Alexander~F Vom~Stein, Stephanie~T J{\"u}nger, Marco Timmer, Ka-Won Noh, Reinhard Buettner, Hamid Kashkar, Volker Neuschmelting, Roland Goldbrunner, and Phuong-Hien Nguyen. 2025.
\newblock The immune landscape in brain metastasis.
\newblock \emph{Neuro-Oncology}, 27(1):50--62.

\bibitem[{Shaaban et~al.(2024)Shaaban, Salem, and Mahmoud}]{shaaban2024cutting}
Aya~M Shaaban, Nancy~M Salem, and Lamees~N Mahmoud. 2024.
\newblock Cutting-edge approaches to cell segmentation in imaging mass cytometry: A detailed review.
\newblock In \emph{2024 6th Novel Intelligent and Leading Emerging Sciences Conference (NILES)}, pages 469--474. IEEE.

\bibitem[{Stanley et~al.(2020)Stanley, Stelzer, Tsai, Fallahzadeh, Ganio, Becker, Phongpreecha, Nassar, Ghaemi, Maric et~al.}]{stanley2020vopo}
Natalie Stanley, Ina~A Stelzer, Amy~S Tsai, Ramin Fallahzadeh, Edward Ganio, Martin Becker, Thanaphong Phongpreecha, Huda Nassar, Sajjad Ghaemi, Ivana Maric, et~al. 2020.
\newblock Vopo leverages cellular heterogeneity for predictive modeling of single-cell data.
\newblock \emph{Nature communications}, 11(1):3738.

\bibitem[{Stringer et~al.(2021)Stringer, Wang, Michaelos, and Pachitariu}]{stringer2021cellpose}
Carsen Stringer, Tim Wang, Michalis Michaelos, and Marius Pachitariu. 2021.
\newblock Cellpose: a generalist algorithm for cellular segmentation.
\newblock \emph{Nature methods}, 18(1):100--106.

\bibitem[{Team et~al.(2024)Team, Riviere, Pathak, Sessa, Hardin, Bhupatiraju, Hussenot, Mesnard, Shahriari, Ram{\'e} et~al.}]{team2024gemma}
Gemma Team, Morgane Riviere, Shreya Pathak, Pier~Giuseppe Sessa, Cassidy Hardin, Surya Bhupatiraju, L{\'e}onard Hussenot, Thomas Mesnard, Bobak Shahriari, Alexandre Ram{\'e}, et~al. 2024.
\newblock Gemma 2: Improving open language models at a practical size.
\newblock \emph{arXiv preprint arXiv:2408.00118}.

\bibitem[{Theodoris et~al.(2023)Theodoris, Xiao, Chopra, Chaffin, Al~Sayed, Hill, Mantineo, Brydon, Zeng, Liu et~al.}]{theodoris2023transfer}
Christina~V Theodoris, Ling Xiao, Anant Chopra, Mark~D Chaffin, Zeina~R Al~Sayed, Matthew~C Hill, Helene Mantineo, Elizabeth~M Brydon, Zexian Zeng, X~Shirley Liu, et~al. 2023.
\newblock Transfer learning enables predictions in network biology.
\newblock \emph{Nature}, 618(7965):616--624.

\bibitem[{Tian et~al.(2023)Tian, Chen, and Macosko}]{tian2023expanding}
Luyi Tian, Fei Chen, and Evan~Z Macosko. 2023.
\newblock The expanding vistas of spatial transcriptomics.
\newblock \emph{Nature Biotechnology}, 41(6):773--782.

\bibitem[{Touvron et~al.(2023)Touvron, Martin, Stone, Albert, Almahairi, Babaei, Bashlykov, Batra, Bhargava, Bhosale et~al.}]{touvron2023llama}
Hugo Touvron, Louis Martin, Kevin Stone, Peter Albert, Amjad Almahairi, Yasmine Babaei, Nikolay Bashlykov, Soumya Batra, Prajjwal Bhargava, Shruti Bhosale, et~al. 2023.
\newblock Llama 2: Open foundation and fine-tuned chat models.
\newblock \emph{arXiv preprint arXiv:2307.09288}.

\bibitem[{Walsh and Quail(2023)}]{walsh2023decoding}
Logan~A Walsh and Daniela~F Quail. 2023.
\newblock Decoding the tumor microenvironment with spatial technologies.
\newblock \emph{Nature Immunology}, 24(12):1982--1993.

\bibitem[{Yun et~al.(2024)Yun, Peng, Trevino, Park, and Chen}]{yun2024mew}
Sukwon Yun, Jie Peng, Alexandro~E Trevino, Chanyoung Park, and Tianlong Chen. 2024.
\newblock Mew: Multiplexed immunofluorescence image analysis through an efficient multiplex network.
\newblock In \emph{European Conference on Computer Vision}, pages 127--144. Springer.

\bibitem[{Zhao et~al.(2024)Zhao, Zhang, Wu, Luo, and Nie}]{zhao2024langcell}
Suyuan Zhao, Jiahuan Zhang, Yushuai Wu, Yizhen Luo, and Zaiqing Nie. 2024.
\newblock Langcell: Language-cell pre-training for cell identity understanding.
\newblock \emph{arXiv preprint arXiv:2405.06708}.

\end{thebibliography}

\appendix




\end{document}